\documentclass[letterpaper, 10 pt, conference]{ieeeconf}  % Comment this line out if you need a4paper

\IEEEoverridecommandlockouts                              % This command is only needed if 
\usepackage[acronym,toc]{glossaries}
\usepackage{graphics}
\usepackage{wrapfig}

\usepackage[style=ieee,hyperref,natbib=true,backend=bibtex,firstinits,doi=false,%
     mincitenames=1,maxcitenames=2,maxbibnames=99,sorting=none,terseinits=false,hyperref=true]{biblatex}
\bibliography{references.bib}
\renewbibmacro*{bbx:savehash}{}% no dash for same authors
\defbibheading{bibliography}[\bibname]{\section*{References}}

\usepackage [T1]{fontenc}
\usepackage[utf8]{inputenc}
\usepackage[table]{xcolor}
\usepackage{hyperref}
\usepackage{listings}
\usepackage{algorithm}
\usepackage{algpseudocode}
\usepackage[acronym,toc]{glossaries}
\usepackage{caption}
\usepackage{mathtools}
\usepackage{todonotes}
\usepackage{dirtytalk}
\usepackage{nicefrac}
\usepackage{multirow}
\usepackage{textcomp}
\usepackage{booktabs}
\usepackage{textcomp}
\usepackage{graphicx}
\usepackage{latexsym}
\usepackage{relsize}
\usepackage{balance}
\usepackage{comment}
\usepackage{xfrac}
\usepackage{enumerate}

\usepackage{dsfont}
\usepackage{graphicx}
\usepackage{color}

\usepackage{fancyvrb}

\usepackage{xcolor}

\lstdefinestyle{base}{
	breaklines=true,
	columns=flexible,
	basicstyle=\footnotesize\ttfamily\color{black},
	moredelim=**[is][\color{red}]{@}{@},
	moredelim=**[is][\color{blue}]{~}{~},
}

\newacronym{LLM}{LLM}{Large Language Model}
\newacronym{CoT}{CoT}{Chain-of-Thought}

\title{\LARGE \bf
A Comparison of Prompt Engineering Techniques\\for Task Planning and Execution in Service Robotics
}

\author{Jonas Bode, Bastian P\"atzold, Raphael Memmesheimer, and Sven Behnke% <-this % stops a space
\thanks{All authors are with the Autonomous Intelligent Systems group, Computer Science Institute VI -- Intelligent Systems and Robotics, Lamarr Institute for Machine Learning and Artificial Intelligence, and Center for Robotics, University of Bonn, Germany; {\tt s6jobode@uni-bonn.de}}
\thanks{This work has been funded by the German Ministry of Education and Research under the grant no. 16SV8683, project: Transferzentrum Roboter im Alltag (RimA).}
}

\begin{document}

\maketitle
\thispagestyle{empty}
\pagestyle{empty}

%%%%%%%%%%%%%%%%%%%%%%%%%%%%%%%%%%%%%%%%%%%%%%%%%%%%%%%%%%%%%%%%%%%%%%%%%%%%%%%%
\begin{abstract}

Recent advances in \glspl{LLM} have been instrumental in autonomous robot control and human-robot interaction by leveraging their vast general knowledge and capabilities to understand and reason across a wide range of tasks and scenarios. Previous works have investigated various prompt engineering techniques for improving the performance of \glspl{LLM} to accomplish tasks, while others have proposed methods that utilize LLMs to plan and execute tasks based on the available functionalities of a given robot platform. In this work, we consider both lines of research by comparing prompt engineering techniques and combinations thereof within the application of high-level task planning and execution in service robotics. We define a diverse set of tasks and a simple set of functionalities in simulation, and measure task completion accuracy and execution time for several state-of-the-art models. We make our code, including all prompts, available at \url{https://github.com/AIS-Bonn/Prompt_Engineering}.

\end{abstract}

%%%%%%%%%%%%%%%%%%%%%%%%%%%%%%%%%%%%%%%%%%%%%%%%%%%%%%%%%%%%%%%%%%%%%%%%%%%%%%%%
\section{Introduction}

Since the unveiling of OpenAI's ChatGPT in November 2022, the rapid emergence of Large Language Models (LLMs) as a major contributor to human-machine interaction has been rippling through society. With advances in scaling, instruction tuning, and alignment, models like GPT~\cite{GPT,Fewshot}, Gemini~\cite{Gemini}, Llama~\cite{Llama2}, or Mistral~\cite{Mistral} can seemingly understand tasks presented in natural language and respond with answers in natural language that are often appropriate. 

In this work, we investigate approaches that leverage these advances for robot control and human-robot interaction~\cite{Microsoft, SayCan, singh2023progprompt, ding2023integrating}, as well as prompt engineering techniques that aim to increase performance across various benchmarks~\cite{Survey, CoT, ReAct}. In particular, we focus on comparing and combining prompt engineering techniques for the application of task planning and execution in the context of service robotics.

Many applications require autonomous robots to perform tasks in complex, cluttered, dynamic, or unknown environments. In order for such a robot to adapt to new tasks and environments, an expert is often required to understand the task logic and translate it into an implementation based on the robot's capabilities. This severely limits speed and flexibility for robot deployment. \glspl{LLM} promise to relax this requirement while allowing for intelligent replanning during task execution. To this end, we focus on the challenging field of service robotics~\cite{StuecklerSB:Cosero16, Memmesheimer:Winner2024}, requiring robots to interact with non-expert operators in open-ended, highly unstructured environments designed for and shared with humans.

\begin{figure}
	\centering
	\includegraphics[width=.77\linewidth]{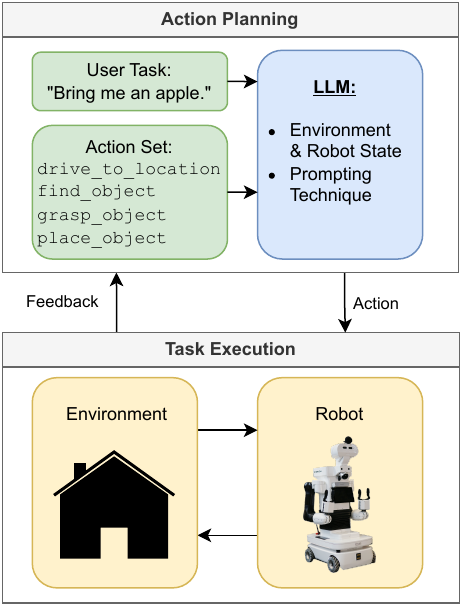}
	\caption{Overview of our simulated experimental setup to evaluate task completion. The user specifies a task in natural language. The action set describes the robot capabilities. Task planning and action selection are performed by prompting an \gls{LLM}. The robot-environment simulation executes the action and provides feedback in form of a changed state.}
	\label{OverviewFig}
    \vspace{-.5cm}
\end{figure}

To use an \gls{LLM} for task planning, one must provide all the necessary information about the environment, the task, and the robot capabilities in textual form. Similarly, for task execution, the implementation of the robot's high-level functionalities must be robust and general enough to account for unknown features of the environment and unexpected events during task execution. In this work, we obviate these problems by employing a simulated environment depicted in Figure~\ref{OverviewFig} to systematically examine the effect of prompt engineering techniques on task completion across several types of tasks and state-of-the-art \glspl{LLM}. 

\noindent The contributions of this paper are as follows:
\begin{itemize}
    \item We derive an \gls{LLM}-based method allowing zero-shot task execution by sampling relevant action sequences from sets of pre-defined actions received via natural language commands from non-expert operators.
    \item We thoroughly investigate and evaluate the feasibility and effectiveness of integrating \glspl{LLM} with various prompt engineering techniques for long-horizon tasks in the domain of service robotics.
\end{itemize}

%We will outline the framework in which we employ the \gls{LLM}, discuss what \gls{LLM} we use, elaborate upon the prompting techniques we choose for our comparisons, and empirically evaluate the effectiveness of different approaches.  

\section{Related Works}
\label{RelWorks}

%In the following, we briefly discuss related work with focus on \glspl{LLM} for robot control and further introduce recent prompting techniques.

Using \glspl{LLM} for high-level robot task planning and for low-level execution control are hot research topics~\cite{abs-2311-07226, abs-2401-04334}.

\subsection{Using \glspl{LLM} for Robot Task Planning \& Execution}

In recent years, \glspl{LLM} have been explored and adopted across a wide range of applications, motivated by harnessing the world knowledge extracted from the vast text data sets they have been trained on.

\textcite{Microsoft} are among the proponents for high-level robot task planning using GPT. They outline the design principles of a robotics pipeline that integrates GPT to plan and implement task execution given an objective in textual form. They define a function library that implements various core functionalities for a given robot platform. They then let GPT generate code that implements a task given by a (non-technical) user based on this function library. Finally, the user can provide feedback and corrections to the code before approving it, allowing the robot to execute it. Their pipeline is evaluated in various domains such as manipulation, aerial navigation, and logical reasoning. This approach can generate complex code that incorporates appropriate branching to handle unexpected events during task execution, provided that the library functions support such information. However, the generated code must explicitly anticipate any such cases beforehand, as the method does not allow for dynamic replanning within GPT.

\textcite{SayCan} present a similar approach where a set of core functionalities is implemented for a given robot platform and exposed to an \gls{LLM} input as context. Instead of relying on human feedback, they assign affordance values to all available functions and execute the highest-ranking one. They recalculate the affordance values before each function call to account for the current state of the robot and the environment.

% \textcite{kannan2023smart} focus on task decomposition and distribution in the context of multi-agent systems.

Another \gls{LLM}-based method for robot task planning is presented by \textcite{singh2023progprompt}, combining strengths in common-sense reasoning and code understanding to generate executable plans. Their experiments demonstrate that incorporating programming language features enhances task performance and adapts well to diverse scenarios.

\textcite{ding2023integrating} propose a method for open-world task planning and situation handling that dynamically integrates commonsense knowledge into a robot's action knowledge, assesses the feasibility of progressing with the existing plan, or determines to adapt the plan accordingly. They integrate their approach on a real robot platform and demonstrate it executing service tasks in a domestic environment.

% \cite{ConCom}

\subsection{Prompt Engineering Techniques}

While \glspl{LLM} are capable of performing tasks with zero-shot prompting~\cite{Oneshot}, they benefit from in-context examples with few-shot prompting~\cite{Fewshot}. To further improve performance on tasks requiring complex reasoning, a variety of prompt engineering techniques have been introduced~\cite{Survey}. In this work, we focus on comparing a selection of relevant and applicable prompt engineering techniques suitable for selecting and dynamically adapting sequences of function calls from a predefined library of functions to accomplish a given task.

\textcite{CoT} introduced \gls{CoT} prompting, which aims to improve performance on complex tasks by incorporating step-by-step reasoning into the responses provided. For natural language responses, this is easily achieved by modifying the original prompt to elicit step-by-step reasoning in the response. In the context of responses that contain function calls, we can first request step-by-step reasoning in a natural language response with a separate prompt before requesting one or multiple appropriate function calls based on that reasoning.

\textcite{ReAct} propose ReAct, which facilitates interleaved planning and action generation. Instead of planning ahead and then executing a complex multi-step plan without further reasoning and adapting to intermediate results, they introduce a discrete reasoning step between each action to ground reasoning in the results of past actions and update action plans accordingly.

%While chat \glspl{LLM} contain a vast amount of knowledge, without retraining they cannot adapt to new knowledge beyond the context of the current dialog. To address this, B\"armann et al.~\cite{Incremental} implement incremental learning. If the robot in this architecture makes a mistake and is corrected by an operator, it can call a dedicated learning function. This summarizes the learned behavior using another \gls{LLM} before adding it to a database. In subsequent conversations, the robot program will then fetch knowledge from this database using a similarity measure to identify relevant knowledge to the current task command. The fetched knowledge is then integrated into the \gls{LLM} prompt. The architecture in \cite{Incremental} is thereby capable of persistent learning and can remember behavior corrections from users. The researchers also noted how the wording of a prompt can significantly impact performance.

\section{Dialogue with a Simulated Environment}
\label{simulation}

We investigate the effect of prompt engineering techniques across multiple tasks in a simulated environment. In this environment, the robot is provided with pre-defined functions from which an \gls{LLM} must select and execute the appropriate ones to accomplish a task given in natural language. This process is modeled as a conversation between the robot and the LLM assistant. Once the \gls{LLM} signals that it has completed the task, we probe the environment to determine if the target condition indicating task success is met. 

\subsection{Environment}

In order to create an environment that allows an accurate comparison of the effect of prompt engineering techniques, we adopt the setting found at RoboCup@Home~\cite{Stueckler:RAM2012,rulebook_2024}, which is a major international competition that focuses on domestic service robots.

The arena is designed to replicate a typical domestic dwelling with several rooms: A study room, a parlor, a kitchen, and a bedroom. 
All tasks start with both robot and operator located in the parlor. 
Each room contains multiple objects, including common food and household items, with which the robot can interact.

\subsection{Action Set}

We use the function-calling feature introduced by OpenAI's API for their GPT line of models~\cite{Fewshot}, which is adopted by other \glspl{LLM} such as Mistral~\cite{Mistral}, as well as several open-source inference pipelines. It allows for specifying the signatures of the pre-defined robot functions as a JSON schema. This exposes the function names, parameters, and parameter types, along with the corresponding natural language descriptions for all functions and parameters, to the context of the \gls{LLM}. The \gls{LLM} can then refer to these functions when generating natural language reasoning or planning, and it can call any of the functions, including the specification of parameter values, as JSON that can be easily parsed and passed to the simulation. The simulation then adjusts the state of the robot and environment accordingly, and returns a natural language text to the \gls{LLM} in response to the function call, providing relevant information about the success of the function and its effect on the robot and environment states.

\vspace*{.5ex}
\noindent We define the same set of five functions for all tasks:

\begin{itemize}
	\item \texttt{drive\_to\_location(location\_name):} This function moves the robot to the specified location. The available options are the exact names of the four rooms in the simulation: \textit{study}, \textit{parlor}, \textit{kitchen}, and \textit{bedroom}.
	\item \texttt{find\_object(object\_name\_list):} The argument of this function is a list of strings that refer to object names the robot will search for in the room it is in. These object names must exactly match the object names specified for each room in the simulation. For each of the objects, the function returns the number of instances found in the room.
	\item \texttt{grasp\_object(object\_name):} Once an object has been found, this function lets the robot grasp the specified object. Again, the name of the object must exactly match an object available in the current room. As an additional constraint, the robot can never carry more than two objects at the same time.
	\item \texttt{place\_object(object\_name):} This function makes the robot place the specified object that is being carried in the current room that the robot is in.
    \item \texttt{exit():} A call to this function signals that the task objective has been accomplished and gracefully terminates the simulation run to the target condition.
\end{itemize}

If the \gls{LLM} returns an invalid function call with respect to the specified JSON schema, the simulation run is considered a failure. If the \gls{LLM} specifies a function parameter that is invalid with respect to the function being unable to execute its intended objective, the function returns a response briefly explaining the problem so that the \gls{LLM} can adapt and continue the simulation run. Each simulation run is automatically terminated and tested for meeting the target condition after 40 function calls have been executed.

\subsection{Tasks}

\begin{figure*}[htpb]
	\centering
	\includegraphics[width=\textwidth]{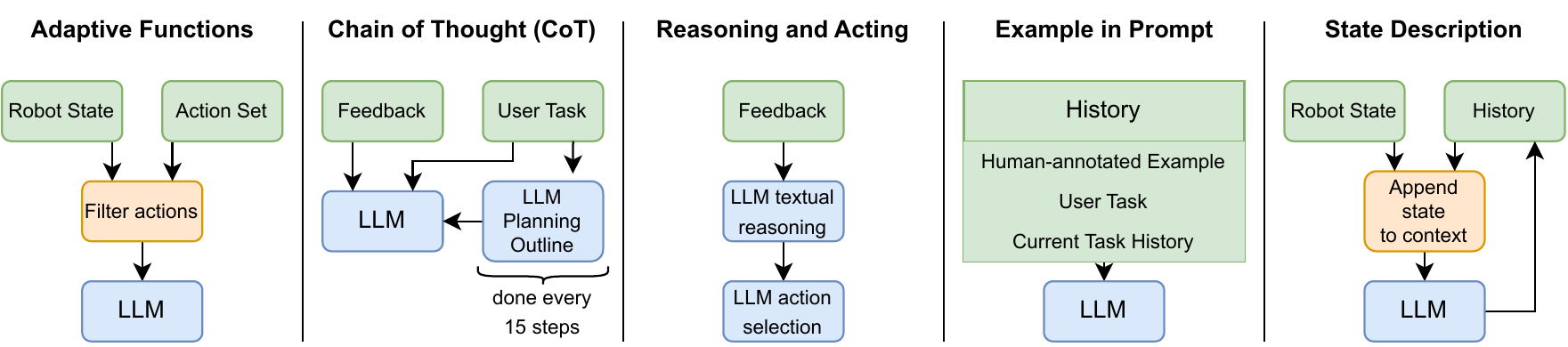}
    \caption{Control flow diagrams of the five prompt engineering techniques examined. See Sec.~\ref{sec:PE} for description.}
	\label{PromptingFig}
\end{figure*}

In our experiments, we define distinct tasks with varying levels of complexity aimed at testing different types of abilities. All tasks are designed to resemble realistic use cases in the sense that they demonstrate a useful application when implemented with a real robot platform, and are related to tasks found in the RoboCup@Home~\cite{rulebook_2024} competition. To demonstrate robustness and generalization, we randomly assign the objects and their respective locations for each task.

\begin{table*}[bthp]
	\begin{center}
		% \caption{Mean time it took to generate one experiment and the success rate for each combination of prompt engineering technique and task. The \gls{LLM} used was GPT\,3.5\,Turbo 0125. Each combination was evaluated with a sample size of $n=20$ and a model temperature of $\tau = 1$.}
		\caption{Comparison of prompt engineering techniques using GPT-3.5-Turbo-0125.}
		\label{tab:results3}
		%\resizebox{\textwidth}{!}{
			%	\centering
                \small
			\begin{tabular}{l|rr|rr|rr|rr}
				% \begin{tabular}{lrrrrrrrr}
					\toprule
					\multirow{2}{*}{\textbf{\begin{tabular}[c]{@{}l@{}}Prompting Technique\end{tabular}}} & \multicolumn{2}{c|}{\textbf{Fetch}}                           & \multicolumn{2}{c|}{\textbf{Conditional}}               & \multicolumn{2}{c|}{\textbf{Equals}}  & \multicolumn{2}{c}{\textbf{Distribute}}                    \\ \cline{2-9} 
					& \multicolumn{1}{c|}{success rate} & \multicolumn{1}{c|}{mean time\,[s]} & \multicolumn{1}{c|}{success} & \multicolumn{1}{c|}{time\,[s]} & \multicolumn{1}{c|}{success} & \multicolumn{1}{c|}{time\,[s]} & \multicolumn{1}{c|}{success} & \multicolumn{1}{c}{time\,[s]} \\ 
					\midrule
					\textbf{Baseline} & \multicolumn{1}{r|}{0.02} & 6.55 & \multicolumn{1}{r|}{0.00} & 9.07 & \multicolumn{1}{r|}{0.00} & 7.83 & \multicolumn{1}{r|}{0.00} & 10.47 \\ %\hline
					\textbf{AF} & \multicolumn{1}{r|}{0.22} & 3.99 & \multicolumn{1}{r|}{0.00} & 6.41 & \multicolumn{1}{r|}{0.00} & 5.23 & \multicolumn{1}{r|}{0.04} & 11.99 \\ %\hline
					\textbf{AF + EiP} & \multicolumn{1}{r|}{\textbf{1.00}} & 5.57 & \multicolumn{1}{r|}{0.38} & 5.43 & \multicolumn{1}{r|}{\textbf{0.20}} & 9.45 & \multicolumn{1}{r|}{0.02} & 13.79 \\ %\hline
					\textbf{AF + CoT} & \multicolumn{1}{r|}{0.46} & 5.66 & \multicolumn{1}{r|}{0.02} & 5.61 & \multicolumn{1}{r|}{0.00} & 6.65 & \multicolumn{1}{r|}{0.00} & 8.25 \\ %\hline
					\textbf{AF + CoT + EiP} & \multicolumn{1}{r|}{\textbf{1.00}} & 6.76 & \multicolumn{1}{r|}{0.44} & 7.45 & \multicolumn{1}{r|}{0.16} & 11.47 & \multicolumn{1}{r|}{0.02} & 17.46 \\ %\hline
					\textbf{AF + ReAct + EiP} & \multicolumn{1}{r|}{\textbf{1.00}} & 14.88 & \multicolumn{1}{r|}{0.44} & 13.76 & \multicolumn{1}{r|}{0.10} & 22.67 & \multicolumn{1}{r|}{\textbf{0.10}} & 25.86 \\ %\hline
					\textbf{AF + StD} & \multicolumn{1}{r|}{0.82} & 4.93 & \multicolumn{1}{r|}{0.00} & 10.12 & \multicolumn{1}{r|}{0.00} & 8.96 & \multicolumn{1}{r|}{0.00} & 9.58 \\ %\hline
					\textbf{AF + CoT + EiP + StD} & \multicolumn{1}{r|}{\textbf{1.00}} & 8.93 & \multicolumn{1}{r|}{0.36} & 14.03 & \multicolumn{1}{r|}{0.00} & 16.40 & \multicolumn{1}{r|}{0.00} & 20.85 \\ %\hline
					\textbf{AF + ReAct + EiP + StD} & \multicolumn{1}{r|}{\textbf{1.00}} & 20.69 & \multicolumn{1}{r|}{\textbf{0.72}} & 41.71 & \multicolumn{1}{r|}{0.14} & 28.37 & \multicolumn{1}{r|}{0.00} & 68.19 \\ %\hline
					\bottomrule
				\end{tabular}\\
				\vspace{0.2cm}
				AF -- prompting with adaptive functions, EiP -- example in prompt, StD -- prompt with appended state description. See Sec.~\ref{sec:PE}.
			\end{center}
			\vspace{-0.7cm}
		\end{table*}

\noindent We define the following four tasks:

\subsubsection{Fetch}

The Fetch task requires the robot to pick up an object from another location and to return it to the user. Since Fetch can be solved with a simple sequence of function calls, it mainly tests whether the \gls{LLM}, in combination with the applied prompt engineering technique, is able to comprehend the scenario, use the function-calling feature, and generate valid JSON appropriate to solve the task. An example of Fetch is: "Please get me a pen from the study". 

\subsubsection{Conditional}

The Conditional task extends Fetch by adding the requirement to gather knowledge about the environment during task execution and to adapt its behavior accordingly. In particular, the robot is asked to search for an object and return one of two other objects to the user. The choice of the latter depends on the first object found. This task therefore tests whether the \gls{LLM} is capable of branching logic. An example of Conditional is: "Check if there is a spoon in the kitchen. If you find one, bring me a pen from the study. If not, bring me a comb from the bedroom".

\subsubsection{Equals}

The Equals task requires the robot to make a numerical observation and repeatedly retrieve an object according to that observation. It therefore tests basic mathematical ability and requires the execution of a logical loop. This results in a long task that requires many function calls, challenging the \gls{LLM} to maintain focus on accomplishing the task objective. An example of Equals is: "For every orange in the kitchen, move a fork from the kitchen to the parlor".

\subsubsection{Distribute}

In the Distribute task, a given object must be distributed so that every location in the simulation contains at least one instance. The \gls{LLM} must therefore not only visit all locations, but also make numerous decisions regarding the movement of objects and keep track of their locations. An example of Distribute is: "Please distribute the pens evenly so that each location contains at least one pen. You can start with the pens in the study".

\section{Experiments}
\label{experiments}

\subsection{\gls{LLM} Variants}

We focus on evaluating GPT~\cite{Fewshot} model variants, due to the function-calling feature of the corresponding API, and their top-of-the-line results on benchmarks regarding this feature~\cite{Leaderboard}. In particular, we use GPT-3.5-Turbo-0125 and GPT-4-0125-Preview, referred to as GPT\,4\,Turbo. 
We evaluate each condition with respect to tasks and prompt engineering techniques with 50 repetitions for GPT\,3.5\,Turbo and 20 repetitions for GPT\,4\,Turbo due to budgetary considerations. 
For both model variants and all tasks, we use a temperature setting of $\tau = 0$ to elicit factual responses. All other model parameters are set to the default values suggested by OpenAI.

\subsection{Prompt Engineering Techniques}
\label{sec:PE}
Our simulation procedure allows the robot to execute tasks immediately after providing the \gls{LLM} with the task description along with the function definitions in an initial prompt. Task completion then proceeds as a dialog of function calls generated by the \gls{LLM} and corresponding function responses generated by the simulation. This approach serves as a baseline for investigating and comparing the effects of various prompt engineering techniques on task completion and time required. We focus on a selection of popular and applicable prompt engineering techniques for task planning and execution in the service robotics domain. Figure~\ref{PromptingFig} gives an overview of this selection.

We investigate (combinations of) the following five prompt engineering techniques:

\subsubsection{Adaptive Functions}

Based on the robot's state and its knowledge of the environment, we can deduce that certain functions cannot accomplish their intended objective. For example, the robot cannot place an object if it is not currently carrying at least one. Similarly, the robot cannot grasp an object if it is already carrying the specified maximum capacity of two objects. Since the tool-calling feature of the targeted API provides the capability to specify the set of available functions with each prompt, we can exclude such functions accordingly. This approach effectively prevents the \gls{LLM} from calling those functions. While this is a useful feature for guard-railing the system in general, we intend to use it to reduce the number of failed function calls, and instead focus the \gls{LLM}'s attention on viable functions. We refer to this technique as Adaptive Functions.

\subsubsection{Chain-of-Thought (CoT)}

We integrate \gls{CoT}~\cite{CoT} prompting into our framework by asking the \gls{LLM} to first provide a textual step-by-step plan detailing how it intends to complete the task given the available functions from the function library. Only after such a plan has been generated, we ask the \gls{LLM} to call the appropriate functions based on that plan. If the task execution turns out to require many steps, we repeat this planning step every 15 function calls, to ensure that the \gls{LLM} is always operating according to a plan in case the initial plan is not detailed enough or incomplete.

\subsubsection{Reasoning and Acting (ReAct)} 

While \gls{CoT} can be characterized as planning a task in advance, likely involving multiple steps, and then executing the plan without explicitly adapting the original plan or any further reasoning.  This may not be sufficient when unexpected or difficult-to-anticipate events occur during task execution. To address this problem, 
we use ReAct~\cite{ReAct} to maintain a tight coupling between task execution and reasoning, by explicitly enforcing the \gls{LLM} to alternate between the two with dedicated prompts. This also has the advantage of making the decisions of the \gls{LLM} easily interpretable by looking at the generated reasoning in relation to the generated behavior. This is a useful tool for manually refining function and parameter descriptions.

\subsubsection{Example in Prompt} 

Guiding the \gls{LLM} with a human-annotated example prior to actual task planning and execution, as suggested by \textcite{CoT} and \textcite{ReAct}, helps to remove ambiguity regarding the interpretation of task and function descriptions, and is often combined with other prompting techniques. Specifically, we prepend a human-expert generated example of a successful Equals task to the context of the \gls{LLM}, including the task description and all function calls and responses. The example is designed to be different from any evaluated experimental task.

\begin{table*}[htbp]
	\begin{center}
		% \caption{Mean time it took to generate one experiment and the success rate for each combination of prompt engineering technique and task. The \gls{LLM} used was GPT\,3.5\,Turbo 0125. Each combination was evaluated with a sample size of $n=20$ and a model temperature of $\tau = 1$.}
		\caption{Comparison of prompt engineering techniques using GPT-4-0125-Preview.}
		\label{tab:results4}
		%\resizebox{\textwidth}{!}{
			%	\centering
                \small
			\begin{tabular}{l|rr|rr|rr|rr}
				% \begin{tabular}{lrrrrrrrr}
					\toprule
					\multirow{2}{*}{\textbf{\begin{tabular}[c]{@{}l@{}}Prompting Technique\end{tabular}}} & \multicolumn{2}{c|}{\textbf{Fetch}}                           & \multicolumn{2}{c|}{\textbf{Conditional}}               & \multicolumn{2}{c|}{\textbf{Equals}}  & \multicolumn{2}{c}{\textbf{Distribute}}                    \\ \cline{2-9} 
					& \multicolumn{1}{c|}{success rate} & \multicolumn{1}{c|}{mean time\,[s]} & \multicolumn{1}{c|}{success} & \multicolumn{1}{c|}{ time\,[s]} & \multicolumn{1}{c|}{success} & \multicolumn{1}{c|}{time\,[s]} & \multicolumn{1}{c|}{success} & \multicolumn{1}{c}{time\,[s]} \\ 
					\midrule
					\textbf{Baseline} & \multicolumn{1}{r|}{\textbf{1.00}} & 11.57 & \multicolumn{1}{r|}{0.90} & 14.72 & \multicolumn{1}{r|}{0.85} & 23.65 & \multicolumn{1}{r|}{0.45} & 37.38 \\ %\hline
					\textbf{AF} & \multicolumn{1}{r|}{\textbf{1.00}} & 10.69 & \multicolumn{1}{r|}{0.65} & 12.06 & \multicolumn{1}{r|}{0.60} & 22.68 & \multicolumn{1}{r|}{\textbf{0.90}} & 33.64 \\ %\hline
					\textbf{AF + EiP} & \multicolumn{1}{r|}{\textbf{1.00}} & 11.21 & \multicolumn{1}{r|}{\textbf{1.00}} & 16.03 & \multicolumn{1}{r|}{0.70} & 21.57 & \multicolumn{1}{r|}{0.65} & 31.97 \\ %\hline
					\textbf{AF + CoT} & \multicolumn{1}{r|}{\textbf{1.00}} & 15.14 & \multicolumn{1}{r|}{0.70} & 18.95 & \multicolumn{1}{r|}{0.70} & 33.03 & \multicolumn{1}{r|}{0.20} & 42.32 \\ %\hline
					\textbf{AF + CoT + EiP} & \multicolumn{1}{r|}{\textbf{1.00}} & 15.07 & \multicolumn{1}{r|}{\textbf{1.00}} & 20.69 & \multicolumn{1}{r|}{0.65} & 26.26 & \multicolumn{1}{r|}{0.80} & 36.89 \\ %\hline
					\textbf{AF + ReAct + EiP} & \multicolumn{1}{r|}{\textbf{1.00}} & 31.05 & \multicolumn{1}{r|}{\textbf{1.00}} & 31.44 & \multicolumn{1}{r|}{0.95} & 66.45 & \multicolumn{1}{r|}{0.70} & 64.25 \\ %\hline
					\textbf{AF + StD} & \multicolumn{1}{r|}{\textbf{1.00}} & 10.45 & \multicolumn{1}{r|}{0.90} & 12.63 & \multicolumn{1}{r|}{0.65} & 32.05 & \multicolumn{1}{r|}{\textbf{0.90}} & 29.57 \\ %\hline
					\textbf{AF + CoT + EiP + StD} & \multicolumn{1}{r|}{\textbf{1.00}} & 15.52 & \multicolumn{1}{r|}{\textbf{1.00}} & 20.63 & \multicolumn{1}{r|}{0.80} & 28.80 & \multicolumn{1}{r|}{0.75} & 36.44 \\ %\hline
					\textbf{AF + ReAct + EiP + StD} & \multicolumn{1}{r|}{\textbf{1.00}} & 29.38 & \multicolumn{1}{r|}{\textbf{1.00}} & 32.86 & \multicolumn{1}{r|}{\textbf{1.00}} & 72.34 & \multicolumn{1}{r|}{0.50} & 72.25 \\ %\hline
					\bottomrule
				\end{tabular}\\
				\vspace{0.2cm}
				AF -- prompting with adaptive functions, EiP -- example in prompt, StD -- prompt with appended state description. See Sec.~\ref{sec:PE}.
			\end{center}
			\vspace{-0.5cm}
		\end{table*}
  
\subsubsection{State Description}

Finally, we introduce a prompt engineering technique that summarizes the robot's state and knowledge of the environment at the end of the context for before each prompt. Since many tasks involve a long horizon, it can be difficult for the \gls{LLM} to accurately assess this information when it is required to gather and deduce it from the entire context. These summaries, which we refer to as state descriptions, attempt to mitigate this challenge. To reduce the number of tokens the \gls{LLM} is prompted with, the context will only ever contain a single state description at the very end of the context. Thus, previous state descriptions are removed from the context with each prompt.

A state description contains information about known objects and their locations, as perceived by the robot. It also details whether the robot is carrying any objects, if so, which ones, and the location of both the robot and the operator.

\section{Results}
\label{results}

\begin{figure}[thbp]
	%\centering
	\begin{lstlisting}[frame=single,style=base] 
@user:@
> Please bring me a sponge from the kitchen.
~system:~
> Briefly summarize the next steps to accomplish the task in text form.
@assistant:@
> 1. Drive to the kitchen.
2. Search for a sponge in the kitchen.
3. Grasp the sponge.
4. Drive back to the parlor.
5. Hand over the sponge to the user.
6. Call the exit function to complete the interaction.
~system:~
> From now on only use function calls to execute the task.
@assistant:@
> drive_to_location: {"location": "kitchen"}
~tool:~
> You successfully arrived in the new location kitchen.
@assistant:@
> find_object: {"object_name_list": ["sponge"]}
~tool:~
> The following items were found in the kitchen: 3 sponges
@assistant:@
> grasp_object: {"object_name": "sponge"}
~tool:~
> You successfully grasped the object sponge.
@assistant:@
> drive_to_location: {"location": "parlor"}
~tool:~
> You successfully arrived in the new location parlor.
@assistant:@
> place_object: {"object_name": "sponge"}
~tool:~
> You successfully placed the object sponge.
@assistant:@
> exit: {}
	\end{lstlisting}
	\caption{Transcript showing a Fetch task using adaptive functions and \Gls{CoT}.}
	\label{Transcript}
 \vspace{-0.3cm}
\end{figure}

For illustration, Figure~\ref{Transcript} shows a sample transcript of a Fetch task using GPT\,4\,Turbo and CoT prompting. The averaged results of our experiments for all tasks and prompt engineering techniques are shown in Table~\ref{tab:results3} for GPT\,3.5\,Turbo and in Table~\ref{tab:results4} for GPT\,4\,Turbo. For all test conditions, we report the success rate and the average time required to generate a sample. Note that these times only reflect the time spent waiting for LLM responses from the OpenAI API, while the required time of robot actions and simulation feedback can be considered negligible.

On the Fetch task, GPT\,4\,Turbo achieves a perfect success rate for all prompting techniques. In comparison, GPT\,3.5\,Turbo requires more complex prompting techniques to achieve a high success rate. The baseline for GPT\,3.5\,Turbo fails to solve the task in almost all cases. 
Using adaptive functions, \gls{CoT} or state descriptions increases the success rate. A perfect success rate for GPT\,3.5\,Turbo is only achieved by providing an example. Similar results can be observed for the success rate of the Conditional task with GPT\,4\,Turbo. Again, a perfect success rate is only achieved by providing an example. 
For the Conditional task with GPT\,3.5\,Turbo, we observe a significant drop in the success rate when an example is missing. This supports the argument that \glspl{LLM} are few-shot learners~\cite{Fewshot}. However, a single example seems to be sufficient to accomplish the task at hand.

We find that more complex prompt engineering techniques do not necessarily lead to higher success rates. On the Equals and Distribute tasks with GPT\,3.5\,Turbo, the techniques without state descriptions outperform the corresponding technique with state descriptions. However, state descriptions do not always have a negative impact. For the GPT\,3.5\,Turbo Conditional task, the addition of state descriptions (AF + ReAct + EiP + StD) achieves the highest success rate of 72\%. We also observe that using only adaptive functions achieves the best success rate with GPT\,4\,Turbo for the Distribute task, while achieving the worst success rate for the Conditional and Equals tasks. There is also no clear favorite between \gls{CoT} and ReAct, as the success rates vary depending on the task, model variant, and other prompt engineering techniques used in conjunction. 

Due to the additional prompts used in all prompting techniques besides the adaptive functions, the measured task completion time depends on the technique used. In general, we observe that \gls{CoT} and even more so ReAct show a longer runtime due to the relatively large number of tokens generated by the \gls{LLM} for planning and reasoning compared to function calls. Using state descriptions increases the time even more for the same reason. Even though the adaptive functions do not utilize additional prompts, we can see a slight reduction in time compared to the baseline, because the \gls{LLM} is less likely to call functions that result in failures. In a real application, the response time of an \gls{LLM} can significantly increase the execution time of a task, although functions that require considerable time themselves provide leeway to mask this problem.

We have successfully used GPT for task planning and execution in several service robotics tasks. We show that the use of appropriate prompt engineering techniques can effectively improve performance. However, overall performance is significantly dependent on the model variant used.

\section{Conclusion}
\label{conclusion}

In this work, we demonstrated the use of \glspl{LLM} to understand user commands to plan and execute tasks given an action set and an environment state in the context of service robotics. We evaluated several prompt engineering techniques in a simulated environment, across different tasks and models. GPT\,3.5\,Turbo performed well on simpler tasks using advanced prompt engineering techniques, but struggled with complex scenarios. GPT\,4\,Turbo managed to complete complex tasks with mixed reliability, while significantly outperforming GPT\,3.5\,Turbo on simpler tasks.

We found that while \gls{CoT}~\cite{CoT} and ReAct~\cite{ReAct} improve task completion rates, the best performance is achieved when combined with an example of a successful completion of a related task. This provides an opportunity for human expert knowledge within a task domain or about the robot platform to contribute to the tuning of a robot's behavior.

Augmenting robotic systems with the reasoning capabilities of \glspl{LLM} through prompt engineering techniques is a promising area of research. It enables progress toward general-purpose service robots capable of operating in challenging open-ended environments where classical approaches to task planning and execution reach their limits. However, to achieve this goal, work on prompt engineering techniques will remain essential to enable safe and reliable task planning and execution while providing adaptability to environments and robot platforms.

%\begin{thebibliography}{99}

% \bibliographystyle{IEEEtran}
% \bibliography{IEEEabrv,references}
%\end{thebibliography}
% \balance
\printbibliography

@misc{Leaderboard,
  title={Berkeley Function Calling Leaderboard},
  author={Fanjia Yan and Huanzhi Mao and Charlie Cheng-Jie Ji and Tianjun Zhang and Shishir G. Patil and Ion Stoica and Joseph E. Gonzalez},
  howpublished={\url{https://gorilla.cs.berkeley.edu/blogs/8_berkeley_function_calling_leaderboard.html}},
  note = {Accessed: March 13, 2024}
}

@misc{Gemini,
  title={{Gemini API}},
  author={{Google AI}},
  howpublished={\url{https://cloud.google.com/vertex-ai/generative-ai/docs/model-reference/gemini}},
  note = {Accessed: March 15, 2024}
}

@article{abs-2311-07226,
    title={Large Language Models for Robotics: A Survey}, 
    author={Fanlong Zeng and Wensheng Gan and Yongheng Wang and others},
    journal={arXiv:2311.07226},    
    year={2023},
}

@article{abs-2401-04334,
    title={Large Language Models for Robotics: Opportunities, Challenges, and Perspectives}, 
    author={Jiaqi Wang and Zihao Wu and Yiwei Li and Hanqi Jiang and Peng Shu and Enze Shi and Huawen Hu and Chong Ma and others},
    journal={arXiv:2401.04334},
    year={2024},
}

@inproceedings{Fewshot,
    title={Language models are few-shot learners},
    author={Brown, Tom and Mann, Benjamin and Ryder, Nick and Subbiah, Melanie and Kaplan, Jared D and Dhariwal, Prafulla and Neelakantan, Arvind and Shyam, Pranav and Sastry, Girish and Askell, Amanda and others},
    booktitle={Advances in Neural Information Processing Systems (Neur{IPS})},
    volume={33},
    pages={1877--1901},
    year={2020}
}

@article{Oneshot,
    title={Language models are unsupervised multitask learners},
    author={Radford, Alec and Wu, Jeffrey and Child, Rewon and Luan, David and Amodei, Dario and Sutskever, Ilya and others},
    journal={OpenAI blog},
    volume={1},
    number={8},
    pages={9},
    year={2019}
}

@article{Survey,
    title={A Systematic Survey of Prompt Engineering in Large Language Models: Techniques and Applications}, 
    author={Pranab Sahoo and Ayush Kumar Singh and Sriparna Saha and Vinija Jain and Samrat Mondal and Aman Chadha},
    journal={arXiv:2402.07927},
    year={2024}
}

@inproceedings{CoT,
	title={{Chain-of-Thought} prompting elicits reasoning in large language models},
	author={Jason Wei and Xuezhi Wang and Dale Schuurmans and Maarten Bosma and Brian Ichter and Fei Xia and Ed Chi and Quoc Le and Denny Zhou},
	booktitle={Advances in Neural Information Processing Systems (Neur{IPS})},
	volume = {35},
    pages = {24824--24837},
    year={2022}
}

@inproceedings{ReAct,
    title={{ReAct}: Synergizing Reasoning and Acting in Language Models},
    author={Shunyu Yao and Jeffrey Zhao and Dian Yu and Nan Du and Izhak Shafran and Karthik R Narasimhan and Yuan Cao},
    booktitle={International Conference on Learning Representations ({ICLR})},
    year={2023}
}

@techreport{Microsoft,
    author = {Vemprala, Sai and Bonatti, Rogerio and Bucker, Arthur and Kapoor, Ashish},
    title = {Chat{GPT} for Robotics: Design Principles and Model Abilities},
    institution = {Microsoft},
    year = {2023},
    number = {MSR-TR-2023-8},
}

@misc{GPT,
	title = {Chat{GPT}},
	author = {{OpenAI}},
	howpublished = {\url{https://openai.com/blog/chatgpt}},
	note = {Accessed: September 25, 2023}
}

@article{Llama2,
    title={Llama 2: Open foundation and fine-tuned chat models},
    author={Touvron, Hugo and Martin, Louis and Stone, Kevin and Albert, Peter and Almahairi, Amjad and Babaei, Yasmine and Bashlykov, Nikolay and others},
    journal={arXiv:2307.09288},
    year={2023}
}

@misc{rulebook_2024,
author = {Hart, Justin and Moriarty, Alexander and Pasternak, Katarzyna
and Kummert, Johannes and Hawkin, Alina and Hassouna, Vanessa
and Pena Narvaez, Juan Diego and others},
title = {{RoboCup@Home} 2024: Rules and Regulations},
year = {2024},
howpublished = {\url{https://github.com/RoboCupAtHome/RuleBook/releases/tag/2024.1}},
}

@inproceedings{SayCan,
    title = {Do As {I} Can, Not As {I} Say: Grounding Language in Robotic Affordances},
    author =  {Ichter, Brian and Brohan, Anthony and Chebotar, Yevgen and Finn, Chelsea and Hausman, Karol and Herzog, Alexander and Ho, Daniel and Ibarz, Julian and others},
    booktitle = {Conf. on Robot Learning (CoRL)},
    volume = {205},
    pages = {287--318},
    year = {2023},
}

@article{Mistral,
	title={{Mistral 7B}},
	author={Albert Q. Jiang and Alexandre Sablayrolles and Arthur Mensch and Chris Bamford and Devendra Singh Chaplot and Diego de las Casas and Florian Bressand and Gianna Lengyel and Guillaume Lample and Lucile Saulnier and Lélio Renard Lavaud and Marie-Anne Lachaux and Pierre Stock and others},
	journal={arXiv:2310.06825},
	year={2023}
}

@inproceedings{singh2023progprompt,
    title={{ProgPrompt}: Generating situated robot task plans using large language models},
    author={Singh, Ishika and Blukis, Valts and Mousavian, Arsalan and Goyal, Ankit and Xu, Danfei and Tremblay, Jonathan and Fox, Dieter and Thomason, Jesse and Garg, Animesh},
    booktitle={IEEE International Conference on Robotics and Automation (ICRA)},
    pages={11523--11530},
    year={2023}
}

@article{ding2023integrating,
    title={Integrating action knowledge and {LLM}s for task planning and situation handling in open worlds},
    author={Ding, Yan and Zhang, Xiaohan and Amiri, Saeid and Cao, Nieqing and Yang, Hao and Kaminski, Andy and Esselink, Chad and Zhang, Shiqi},
    journal={Auton. Robots},
    volume={47},
    number={8},
    pages={981--997},
    year={2023},
}

@article{StuecklerSB:Cosero16,
  author       = {J{\"{o}}rg St{\"{u}}ckler and
                  Max Schwarz and
                  Sven Behnke},
  title        = {Mobile Manipulation, Tool Use, and Intuitive Interaction for Cognitive
                  Service Robot {Cosero}},
  journal      = {Frontiers Robotics {AI}},
  volume       = {3},
  pages        = {58},
  year         = {2016},
}

@ARTICLE{Stueckler:RAM2012,
  author={St{\"{u}}ckler, J{\"{o}}rg and Holz, Dirk and Behnke, Sven},
  journal={IEEE Robotics and Automation Magazine}, 
  title={{RoboCup@Home}: Demonstrating Everyday Manipulation Skills in {RoboCup@Home}}, 
  year={2012},
  volume={19},
  number={2},
  pages={34-42},
  doi={10.1109/MRA.2012.2191993}
}

@inproceedings{Memmesheimer:Winner2024,
    title={{RoboCup@Home} 2024 OPL Winner NimbRo: Anthropomorphic Service Robots using Foundation Models for Perception and Planning},
    author={Raphael Memmesheimer and Jan Nogga and Bastian Pätzold and Evgenii Kruzhkov and Simon Bultmann and Michael Schreiber and Jonas Bode and Bertan Karacora and Juhui Park and Alena Savinykh and Sven Behnke},
    booktitle={RoboCup 2024: Robot World Cup XXVII},
    publisher={Springer},
    year={2025},
    note={to appear}
}
% \balance

\end{document}